\documentclass[conference]{IEEEtran}
\IEEEoverridecommandlockouts
\usepackage{cite}
\usepackage{amsmath,amssymb,amsfonts}
\usepackage{algorithmic}
\usepackage{graphicx}
\usepackage{textcomp}
\usepackage{xcolor}
\usepackage{multirow}
\def\BibTeX{{\rm B\kern-.05em{\sc i\kern-.025em b}\kern-.08em
    T\kern-.1667em\lower.7ex\hbox{E}\kern-.125emX}}
\begin{document}

\title{Morphing Attack Potential
\thanks{This work is part of the iMARS project. The project received funding from the European Union’s Horizon 2020 research and innovation program under Grant Agreement No. 883356. \\Disclaimer: this text reflects only the author’s views, and the Commission is not liable for any use that may be made of the information contained therein.
\\ \\This paper is a preprint of a paper accepted by IEEE International Workshop on Biometrics and Forensics (IWBF 2022). When the final version is published, the copy of record will be available at the IEEE Xplore.}
}

\makeatletter
\newcommand{\linebreakand}{%
  \end{@IEEEauthorhalign}
  \hfill\mbox{}\par
  \mbox{}\hfill\begin{@IEEEauthorhalign}
}
\makeatother

\author{
\IEEEauthorblockN{Matteo Ferrara}
\IEEEauthorblockA{\textit{Department of Computer Science and Engineering} \\
\textit{University of Bologna}\\
Cesena, Italy \\
matteo.ferrara@unibo.it}
\and
\IEEEauthorblockN{Annalisa Franco}
\IEEEauthorblockA{\textit{Department of Computer Science and Engineering} \\
\textit{University of Bologna}\\
Cesena, Italy \\
annalisa.franco@unibo.it}
\and
\linebreakand
\IEEEauthorblockN{Davide Maltoni}
\IEEEauthorblockA{\textit{Department of Computer Science and Engineering} \\
\textit{University of Bologna}\\
Cesena, Italy \\
davide.maltoni@unibo.it}
\and
\IEEEauthorblockN{Christoph Busch}
\IEEEauthorblockA{\textit{Norwegian Biometrics Laboratory (NBL)} \\
\textit{Norwegian University of Science and Technology (NTNU)}\\
Gjøvik, Norway \\
christoph.busch@ntnu.no}
}

\IEEEoverridecommandlockouts
\IEEEpubid{\makebox[\columnwidth]{978-1-6654-6962-3/22/\$31.00~\copyright2022 IEEE \hfill} \hspace{\columnsep}\makebox[\columnwidth]{ }}

\maketitle

\IEEEpubidadjcol

\begin{abstract}
In security systems the risk assessment in the sense of common criteria testing is a very relevant topic; this requires quantifying the attack potential in terms of the expertise of the attacker, his knowledge about the target and access to equipment. Contrary to those attacks, the recently revealed morphing attacks against Face Recognition Systems (FRSs) can not be assessed by any of the above criteria. But not all morphing techniques pose the same risk for an operational face recognition system. This paper introduces with the Morphing Attack Potential (MAP) a consistent methodology, that can quantify the risk, which a certain morphing attack creates.
\end{abstract}


\section{Introduction}
Biometric face recognition is nowadays widely in use in border control applications for recognition of individuals based on their biological facial characteristics. The face reference image stored in a passport or other types of identity documents shall provide a strong link between the document holder and the biometric reference incorporated in the document. 
Even though the accuracy of biometric systems is constantly increasing, at the same time face recognition systems must remain tolerant against intra-subject variation. This tolerance bound is what has been exploited in recent years by morphing attacks. In consequence a manipulated face image, generated by merging samples of two or multiple subjects in the image domain, might be verified successfully against probe images captured from the contributing subjects. This morphing attack can be executed, for instance, in the passport application process, where in most European countries the applicant brings his own printed photograph. This way, the unique link between individuals and their biometric reference data is compromised. The first morphing attack of this kind was presented in 2014 by Ferrara et al \cite{6996240}. Meanwhile many hand-crafted or deep-learning based morphing attacks have been described. A good overview is given by Scherhag et al\cite{8053499} and Venkatesh et al\cite{VenkateshRRB20-0}.
While the security risk of morphing attacks for operational face recognition systems is obvious, the methodology to quantify the problem is only in its infancy. Unfortunately methodologies of common criteria testing by means of quantifying the attack potential in terms of expertise of the attacker and his knowledge about the target system are not applicable. But nonetheless serious morphing techniques pose a significant risk for an operational face recognition system. This paper introduces with the Morphing Attack Potential (MAP) a new methodology, that can quantify the risk, which a certain morphing attack creates.

\section{Related works}
The vulnerability of face recognition systems is usually measured on specific databases of morphed images and is quantified as the proportion of morphed images that are erroneously accepted as bona fide samples in a series of face verification attempts. Recently, two metrics have been introduced for vulnerability assessment and are commonly used by the scientific community. The rational of each metric will be briefly described here, while a more formal definition will be provided in Section \ref{sec:map}.
\begin{itemize}
    \item \textbf{Mated Morph Presentation Match Rate (MMPMR)}. Initially proposed by Scherhag et al.\cite{8053499}, it defines the proportion of
morphed images verified with all of its contributing subjects. The basic definition of MMPMR considers a single probe image for each subject, but in case of multiple samples, the MMPMR can be generalized as follows:
\begin{itemize}
    \item{\textit{MinMax-MMPMR}}: for smaller number of samples, multiple comparisons can be
interpreted as multiple authentication attempts per subject. Thus the metric is extended by only considering the maximum (for similarity scores) or minimum (for dissimilarity scores) over all mated morph comparisons of one subject. In this case, the rationale is that a morphing attack succeeds if the morphed image can be successfully verified against at least one of the probe images of each subject, meaning that all contributing subjects can use the passport eventhough they may need multiple attempts (within the attempt limit of the operational system).
    \item{\textit{ProdAvg-MMPMR}}: for larger number of probe samples per subject, the \textit{MinMax} approach is prone to falsely increase the number of accepted subjects. Thus, the authors proposed a probabilistic interpretation, by calculating the proportion of accepted attempts per subject and then multiplying the probabilities of all contributing subjects (i.e., joint probability).
\end{itemize}

\item \textbf{Fully Mated Morph Presentation Match Rate (FMMPMR)}\cite{VenkateshRRB20-0}. It defines the proportion of morphed images verified with its contributing subjects under the condition that the morphed image verifies successfully against both contributing subjects, taking into account multiple authentication attempts for each subject. However, while MMPMR considers the attack as successful if at least one of the multiple authentication attempts succeeds for all contributing subjects, FMMPMR is much more restrictive and requires that all attempts for all contributing subjects achieve a similarity (dissimilarity) score higher (lower) than the fixed threshold.
\end{itemize}

These metrics are complemented by those proposed to assess the robustness of Morphing Attack Detection (MAD) algorithms, defined in the International Standard ISO/IEC 30107-3:2017 \cite{ISO_IEC_30107-3_2017}, such as APCER (Attack Presentation Classification Error Rate) and BPCER (Bona fide Presentation Classification Error Rate). It is worth noting that such metrics can be used to quantify the capability of a morphing detection approach to detect an attack but are not suitable to quantify the vulnerability of face recognition systems.

\section{Morphing Attack Potential}
\label{sec:map}
The aim of this work is to extend the existing metrics for assessment of the vulnerability and measurement of an effective attack potential taking into account: i) a variable number of attempts (e.g. multiple probe images acquired in an ABC gate) for each subject and ii) multiple FRSs.  
In particular, this paper introduces a new metric called Morphing Attack Potential (MAP) aimed at quantifying the attack potential of a given dataset $\mathbb{M}$ of morphed images analyzing the combined impact of the two above cited factors. In this section, we will retrace the definition of existing indicators (MMPMR and FMMPMR) to then come to the definition of MAP. In all cases, each indicator $V$ reports the proportion of morphed images in $\mathbb{M}$ for which a given condition $C_V(M)$ is met:
\begin{equation}
    V=\frac{|M \in \mathbb{M}: C_V(M)=true|}{|\mathbb{M}|}
\end{equation}
The specific condition used for each indicator will be described in the following paragraphs.

Given a morphed image $M$ and a probe image $P$, a verification attempt is considered successful for a FRS $F$ if the similarity score $s_F(M,P)$ between the morphed image and the probe is higher than a prefixed threshold $\tau(F)$ (or the dissimilarity score is below a fixed threshold). Without loss of generality we can consider here a formulation based on similarity scores. Further we elaborate in the following the special case of only two contributing data subjects, which attempt to use the passport at different times. However the proposed metric can easily be extended to multiple contributing subjects (see Appendix).

Let's define the function $m(M,P,F)$, which returns the result of the verification attempt, as follows:
\begin{equation}
    m(M,P,F)=
    \begin{cases}
        1,& \text{if } s_F(M,P)>\tau(F)\\
        0,              & \text{otherwise}
    \end{cases}
\end{equation}

The basic MMPMR indicator measures the proportion of morphed images that are successfully verified against both contributing subjects. In particular, given the probe images $P_1$ and $P_2$ of the two subjects, the condition to be verified is as follows:
\begin{equation}
    C_{\textrm{MMPMR}}(M) := \bigwedge
    \begin{aligned}
    m(M,P_1,F)=1 \\ 
    m(M,P_2,F)=1    
    \end{aligned}
\end{equation}
where the $\bigwedge$ operator is used here to represents the logical AND between the conditions in the two rows.

The FMMPMR indicator takes into account a set of probe images $\mathbb{P}=\{P_1,..,P_m\}$ for each subject ($\mathbb{P}_1$ and $\mathbb{P}_2$), representing for instance multiple video frames acquired at the ABC gate, rather than a single probe. Let's define therefore $mc(M,\mathbb{P},F)$ as a function that returns the number of probe images in $\mathbb{P}$ that are successfully verified against the morphed image $M$:
\begin{equation}
    mc(M,\mathbb{P},F)=|P_i\in\mathbb{P}:m(M,P_i,F)=1|
\end{equation}
Considering two contributing subjects, the condition to be evaluated for a morphed image to be included in FMMPMR is the following:

\begin{equation}
    C_{\textrm{FMMPMR}}(M):= \bigwedge 
    \begin{aligned}
       mc(M,\mathbb{P}_1,F)=|\mathbb{P}_1| \\ 
       mc(M,\mathbb{P}_2,F)=|\mathbb{P}_2| 
    \end{aligned}
\end{equation}

With the MAP definition, we want to generalize further, taking into account a variable number of probe images and multiple face recognition systems. In particular, given a dataset of morphed images $\mathbb{M}$, $m$ probe images for each contributing subject and $n$ FRSs to evaluate, MAP is defined as a matrix of size $(m\times n)$ whose generic element $\mbox{MAP}[r,c]$ with rows $r\in {1,..,m}$ and columns $c\in {1,..,n}$ reports the proportion of morphed images that can successfully reach a match decision with both contributing subjects in at least $r$ verification attempts by at least $c$ FRSs. 

Let $\mathbb{F}=\{F_1,..,F_n\}$ be the set of FRSs considered for attack potential assessment.
We can now define a new function $fmc(M,\mathbb{P},\mathbb{F},r)$ that returns the number of FRSs in $\mathbb{F}$ for which at least $r$ of the probe images in $\mathbb{P}$ are successfully verified against the morphed image $M$:
\begin{equation}
    fmc(M,\mathbb{P},\mathbb{F},r)=|F_i\in\mathbb{F}:mc(M,\mathbb{P},F_i)\geq r|
\end{equation}

The generic element of the proposed indicator, $\mbox{MAP}[r,c]$, reports the proportion of morphed images in $\mathbb{M}$ for which $fmc(M,\mathbb{P},\mathbb{F},r)\geq c$ for both contributing subjects. The condition to be verified is therefore:

\begin{equation}
    C_{\textrm{MAP[\textit{r},\textit{c}]}}(M)=\bigwedge
    \begin{aligned}
        fmc(M,\mathbb{P}_1,\mathbb{F},r)\geq c\\
        fmc(M,\mathbb{P}_2,\mathbb{F},r)\geq c
    \end{aligned}
\end{equation}

A visual example of MAP matrix is  provided in Fig.~\ref{fig:MAP_Example} where artificial data have been used to show the information content of MAP, emphasized by the use of a color map. The two dimensions of the MAP matrix can be related to two very important aspects to take into account for attack potential assessment. Indeed, starting from the top-left corner we can evaluate:
\begin{itemize}
    \item \textit{robustness} - moving from the matrix top to the bottom, we analyze the capability of morphed image to be successfully verified against an increasing number of probe images; the images considered in the last rows present therefore a similarity with both contributing subjects which is steady and not ``accidental'' (as in case of match with a single probe). 
    \item \textit{generality} - moving from left to right, we evaluate the capability of the morphed images to fool an increasing number of FRSs, thus meaning that the morphed images included in the last column present a more general attack potential.
\end{itemize}

It is clear that the most "dangerous" images are those in the bottom-left corner since they exhibit a high degree of both robustness and generality. However the number reported in that corner is low, as only a subset will meet both criteria.

\begin{figure}[ht]
    \centering
    \includegraphics[width=0.7\columnwidth]{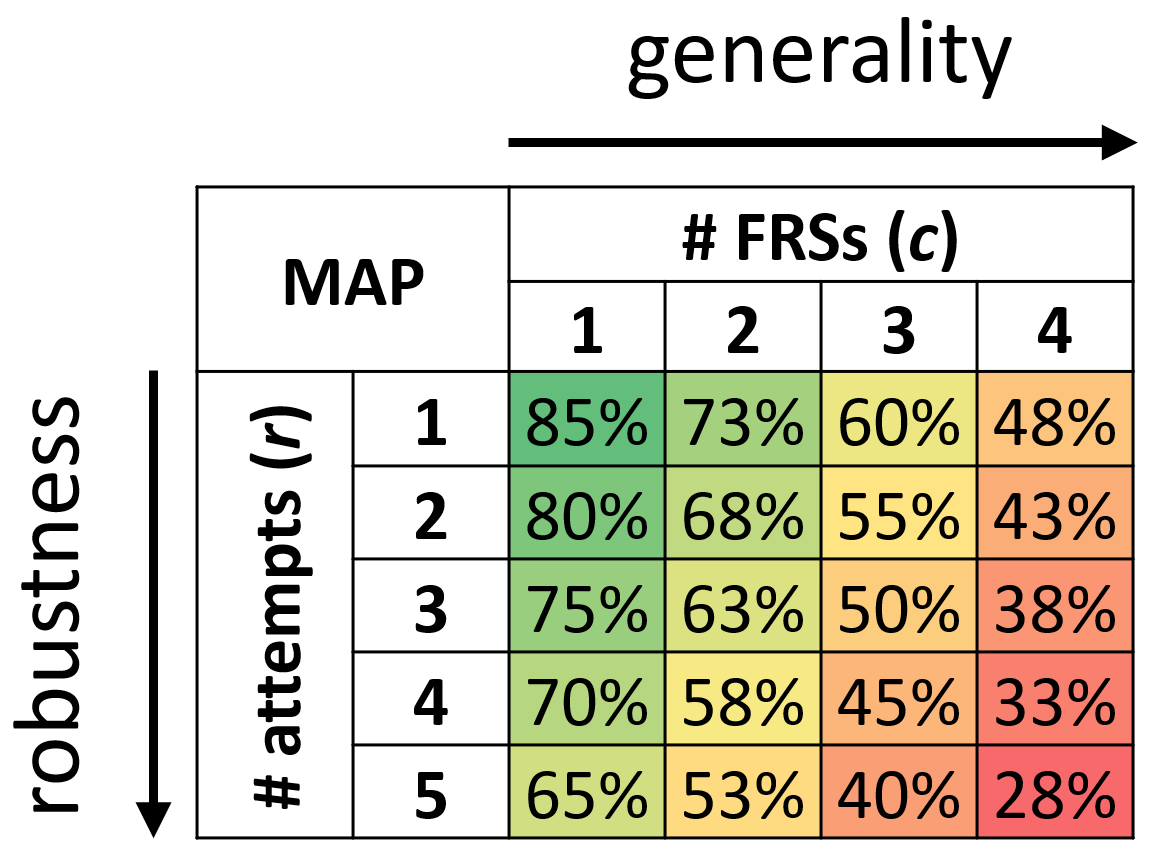}
    \caption{An example of MAP matrix with five rows, referred to 5 verification attempts for each subject, and four columns related to four FRSs used for the evaluation. The generic element $\mbox{MAP}[r,c]$ reports the proportion (expressed as a percentage) of morphed images that successfully reach a match decision against at least $r$ attempts for each contributing subject by at least $c$ of the FRSs evaluated.}
    \label{fig:MAP_Example}
\end{figure}

\section{Experimental results}
This section reports the MAP metric computed on a large database of morphed images \cite{9246583}. Since the main goal of this work is to introduce a new methodology to quantify the risk of morphing attacks, we decided not to use commercial FRSs to avoid disclosure of sensitive information about vulnerabilities of real operating systems installed at ABC gates. We used therefore four state-of-the-art deep models for face recognition ($n=4$) publicly available in the Deepface library \cite{deepface}: ArcFace, Dlib, Facenet and VGG-Face. 

In order to simulate a realistic attack to an ABC system, the operational threshold of each FRS has been fixed according to the Frontex guidelines \cite{FrontexGuidelines}. In particular, for ABC systems operating in verification mode, the face recognition algorithm has to ensure a False Match Rate (FMR) equal to 0.1\% and a False Non-Match Rate (FNMR) lower than 5\%\footnote{According to the International Standard ISO/IEC 19795-1 \cite{ISO_IEC_19795-1} the FAR=FMR(1-FTAR), and the FRR=FTAR+FNMR(1-FTAR), where FTAR is the Failure-To-Acquire Rate. As the MAP is evaluated as technology evaluation on a corpus of morphed images, the FTAR is anticipated as zero. Thus for sake of clarity, the target security level is fixed at a FMR of 0.1\% and a FNMR lower than 5\%.}. During the experimentation, for each FRS, the FMR 0.1\% threshold computed on 19944 non-mated comparison attempts from the Face Morphing Challenge (FMC) database \cite{Ferrara2016} has been used (see Table \ref{table:FRSThr}).

\begin{table}[ht]
\caption{Operational thresholds computed for the four FRSs in order to achieve a FMR=0.1\%}
\begin{center}
\begin{tabular}{|c|c|}
\hline
\textbf{FRS} & \textbf{\textit{Thr}}\\
\hline
ArcFace & 0.49320  \\
\hline
Dlib & 0.04146  \\
\hline
Facenet & 0.26440  \\
\hline
VGG-Face & 0.17400  \\
\hline
\end{tabular}
\end{center}
\label{table:FRSThr}
\end{table}

\subsection{Data set}
The FRSs have been evaluated on the SOTAMD dataset \cite{9246583} containing 5748 high quality morphed images for benchmarking under realistic conditions. The morphed images have been created from facial images of 150 different subjects of various ethnicities, age-groups and both genders. After a careful subject pre-selection, the morphed images have been generated using 7 different morphing algorithms with morphing factor $\alpha=0.3$ and $0.5$, and applying to part of the images a manual post-processing to remove visible artifacts. Moreover, part of the images have been also printed and scanned. To summarize, the database of morphed images used for the evaluation consists of 2045 digital images and 3703 printed/scanned (P\&S). Furthermore, for face verification attempts, a set of probe images of each subject is needed. The SOTAMD dataset also includes ten probe images for each subject ($m=10$), captured under a simulated ABC gate scenario. 

To simulate the verification process at the gate, each morphed image (i.e., contained in the document) has been compared using the four FRSs against all ten gate images (representing multiple video frames) of both subjects involved in the generation of the morphed image. 

\subsection{Results}
The MAP metric for the entire SOTAMD dataset is reported in this section, as well as for different subsets of images as this allows to analyze the impact on the attack potential of some important factors such as the image format (digital vs. P\&S), the morphing factor ($\alpha$) and the morphing algorithm used.

Table~\ref{table:all} reports the MAP values on the whole SOTAMD dataset. Overall, the measured attack potential values are quite low, and only a limited part of the morphed images is able to fool multiple FRSs. This behaviour is mainly due to the FRSs used in our experimentation which do not comply with the Frontex requirements (configured for a maximum FNMR=5\% @ FMR 0.1\%), as confirmed by the high False Non-Match Rates measured on the SOTAMD bona fide images, reported in Table~\ref{table:FRR}. 

\begin{table}[ht]
\begin{center}
\caption{MAP computed on the entire SOTAMD dataset (expressed as a percentage) \\ 5748 images}
\label{table:all}
\begin{tabular}{|c|c|c|c|c|c|}
\cline{3-6}
\multicolumn{2}{c|}{\multirow{2}{*}{}} & \multicolumn{4}{|c|}{\textbf{\# FRSs (\textit{c})}}\\ \cline{3-6}
\multicolumn{2}{c|}{} & \textbf{1} & \textbf{2} & \textbf{3} & \textbf{4}\\
\hline
\multirow{10}{*}{\rotatebox[origin=c]{90}{\textbf{\# Attempts (\textit{r})}}} & \textbf{1} & 39.6\% & 16.8\% & 6.1\% & 1.3\%  \\ \cline{2-6}
& \textbf{2} & 32.9\% & 12.6\% & 4.5\% & 0.9\%  \\ \cline{2-6}
& \textbf{3} & 29.3\% & 10.5\% & 3.5\% & 0.5\%  \\ \cline{2-6}
& \textbf{4} & 26.0\% & 8.4\% & 2.4\% & 0.4\%  \\ \cline{2-6}
& \textbf{5} & 23.4\% & 6.8\% & 1.9\% & 0.2\%  \\ \cline{2-6}
& \textbf{6} & 19.7\% & 5.6\% & 1.4\% & 0.1\%  \\ \cline{2-6}
& \textbf{7} & 16.4\% & 4.6\% & 1.0\% & 0.1\%  \\ \cline{2-6}
& \textbf{8} & 13.7\% & 3.7\% & 0.7\% & 0.1\% \\ \cline{2-6}
& \textbf{9} & 11.5\% & 2.6\% & 0.3\% & 0.0\%  \\ \cline{2-6}
& \textbf{10} & 7.6\% & 1.6\% & 0.0\% & 0.0\%  \\
\hline
\end{tabular}
\end{center}
\end{table}

\begin{table}[ht]
\caption{FNMR measured on the SOTAMD bona fide images using for each FRS the computed FMR=0.1\% threshold (expressed as percentage)}
\begin{center}
\begin{tabular}{|c|c|c|c|}
\hline
\multirow{2}{*}{\textbf{FRS}} & \multicolumn{3}{|c|}{\textbf{FNMR on SOTAMD}} \\ \cline{2-4}
 & \textbf{Entire} & \textbf{Digital} & \textbf{P\&S} \\
\hline
ArcFace & 6.0\% & 5.4\% & 6.2\%  \\
\hline
Dlib & 29.8\% & 28.2\% & 30.2\%  \\
\hline
Facenet & 27.5\% & 24.4\% & 28.4\%  \\
\hline
VGG-Face & 31.4\% & 28.3\% & 32.2\%  \\
\hline
\end{tabular}
\end{center}
\label{table:FRR}
\end{table}

Only ArcFace presents an acceptable FNMR, while the other systems would be inoperable in a real scenario since they tend to reject a large part of the attempts (resulting in an apparent lower potential of the morphing attack). As a general indication, we suggest to ensure that the set $\mathbb{F}$ of FRSs represents complementary recognition methodologies, and contains elements that all have an equivalent recognition performance (compliant with the Frontex requriements).
With reference to the number of attempts, we can conclude that a portion of morphed images analyzed is quite robust, meaning that they can reach a match decision with both contributing subject with up to ten live gate images.

A characteristic of the MAP to point out is that it mainly focuses on the attack potential of a set (or subset) of morphed images rather that on the vulnerability of a specific FRS whose failure cases can be spread over different cells of the matrix; each column of MAP, in fact, is not directly linked to one of the FRSs used for the evaluation.

Tables \ref{table:D} and \ref{table:PS} report respectively MAP for the SOTAMD digital and P\&S images. A comparison between the two matrices (D and P\&S) does not highlight noticeable differences, even if in general the values measured on digital images are slightly higher.

\begin{table}[ht]
\begin{center}
\caption{MAP computed on the SOTAMD subset of digital images (expressed as a percentage) \\ 2045 images}
\label{table:D}
\begin{tabular}{|c|c|c|c|c|c|}
\cline{3-6}
\multicolumn{2}{c|}{\multirow{2}{*}{}} & \multicolumn{4}{|c|}{\textbf{\# FRSs (\textit{c})}}\\ \cline{3-6}
\multicolumn{2}{c|}{} & \textbf{1} & \textbf{2} & \textbf{3} & \textbf{4}\\
\hline
\multirow{10}{*}{\rotatebox[origin=c]{90}{\textbf{\# Attempts (\textit{r})}}} & \textbf{1} & 40.5\% & 17.6\% & 6.8\% & 1.4\%  \\ \cline{2-6}
& \textbf{2} & 33.4\% & 13.3\% & 4.7\% & 0.8\%  \\ \cline{2-6}
& \textbf{3} & 30.3\% & 11.1\% & 3.7\% & 0.6\%  \\ \cline{2-6}
& \textbf{4} & 27.1\% & 8.9\% & 2.7\% & 0.3\%  \\ \cline{2-6}
& \textbf{5} & 24.4\% & 7.0\% & 2.2\% & 0.2\%  \\ \cline{2-6}
& \textbf{6} & 20.2\% & 5.9\% & 1.6\% & 0.0\%  \\ \cline{2-6}
& \textbf{7} & 17.0\% & 5.0\% & 1.2\% & 0.0\%  \\ \cline{2-6}
& \textbf{8} & 14.6\% & 4.2\% & 1.0\% & 0.0\% \\ \cline{2-6}
& \textbf{9} & 12.3\% & 3.0\% & 0.3\% & 0.0\%  \\ \cline{2-6}
& \textbf{10} & 8.5\% & 2.2\% & 0.0\% & 0.0\%  \\
\hline
\end{tabular}
\end{center}
\end{table}

\begin{table}[ht]
\begin{center}
\caption{MAP computed on the SOTAMD subset of printed/scanned images (expressed as a percentage) \\ 3703 images}
\label{table:PS}
\begin{tabular}{|c|c|c|c|c|c|}
\cline{3-6}
\multicolumn{2}{c|}{\multirow{2}{*}{}} & \multicolumn{4}{|c|}{\textbf{\# FRSs (\textit{c})}}\\ \cline{3-6}
\multicolumn{2}{c|}{} & \textbf{1} & \textbf{2} & \textbf{3} & \textbf{4}\\
\hline
\multirow{10}{*}{\rotatebox[origin=c]{90}{\textbf{\# Attempts (\textit{r})}}} & \textbf{1} & 39.1\% & 16.3\% & 5.7\% & 1.3\%  \\ \cline{2-6}
& \textbf{2} & 32.6\% & 12.3\% & 4.3\% & 0.9\%  \\ \cline{2-6}
& \textbf{3} & 28.7\% & 10.2\% & 3.5\% & 0.5\%  \\ \cline{2-6}
& \textbf{4} & 25.5\% & 8.1\% & 2.3\% & 0.4\%  \\ \cline{2-6}
& \textbf{5} & 22.8\% & 6.7\% & 1.8\% & 0.3\%  \\ \cline{2-6}
& \textbf{6} & 19.5\% & 5.4\% & 1.3\% & 0.2\%  \\ \cline{2-6}
& \textbf{7} & 16.1\% & 4.4\% & 0.9\% & 0.1\%  \\ \cline{2-6}
& \textbf{8} & 13.3\% & 3.5\% & 0.6\% & 0.1\%  \\ \cline{2-6}
& \textbf{9} & 11.1\% & 2.4\% & 0.2\% & 0.0\%  \\ \cline{2-6}
& \textbf{10} & 7.0\% & 1.2\% & 0.0\% & 0.0\%  \\
\hline
\end{tabular}
\end{center}
\end{table}

The impact of the morphing factor $\alpha$ used for morphing generation (0.5 or 0.3) can be evaluated by comparing Tables \ref{table:alfa05} and \ref{table:alfa03}. As expected, in this case the difference is quite clear, morphing factor $\alpha=0.5$ produces morphed images whose chance of success is about twice that of images generated with $\alpha=0.3$ for all the MAP cells. This is because morphing factor $\alpha=0.5$ is the best choice when the objective is to generate morphed images with the maximum probability to be successfully verified against probe images of both contributing subjects. This finding validates the applicability of the MAP as a holistic metric.

\begin{table}[ht]
\begin{center}
\caption{MAP computed on the SOTAMD subset of images with $\alpha=0.5$ (expressed as a percentage) \\ 2860 images}
\label{table:alfa05}
\begin{tabular}{|c|c|c|c|c|c|}
\cline{3-6}
\multicolumn{2}{c|}{\multirow{2}{*}{}} & \multicolumn{4}{|c|}{\textbf{\# FRSs (\textit{c})}}\\ \cline{3-6}
\multicolumn{2}{c|}{} & \textbf{1} & \textbf{2} & \textbf{3} & \textbf{4}\\
\hline
\multirow{10}{*}{\rotatebox[origin=c]{90}{\textbf{\# Attempts (\textit{r})}}} & \textbf{1} & 53.5\% & 25.2\% & 9.5\% & 1.8\%  \\ \cline{2-6}
& \textbf{2} & 45.2\% & 19.3\% & 6.9\% & 1.2\%  \\ \cline{2-6}
& \textbf{3} & 40.8\% & 16.2\% & 5.5\% & 0.7\%  \\ \cline{2-6}
& \textbf{4} & 36.5\% & 12.8\% & 3.4\% & 0.4\%  \\ \cline{2-6}
& \textbf{5} & 33.0\% & 9.9\% & 2.5\% & 0.3\%  \\ \cline{2-6}
& \textbf{6} & 28.3\% & 7.8\% & 1.8\% & 0.2\%  \\ \cline{2-6}
& \textbf{7} & 23.4\% & 6.3\% & 1.3\% & 0.1\%  \\ \cline{2-6}
& \textbf{8} & 19.5\% & 4.9\% & 0.9\% & 0.1\% \\ \cline{2-6}
& \textbf{9} & 16.1\% & 3.4\% & 0.3\% & 0.0\% \\ \cline{2-6}
& \textbf{10} & 10.3\% & 2.2\% & 0.0\% & 0.0\% \\
\hline
\end{tabular}
\end{center}
\end{table}

\begin{table}[ht]
\begin{center}
\caption{MAP computed on the SOTAMD subset of images with $\alpha=0.3$ (expressed as a percentage) \\ 2888 images}
\label{table:alfa03}
\begin{tabular}{|c|c|c|c|c|c|}
\cline{3-6}
\multicolumn{2}{c|}{\multirow{2}{*}{}} & \multicolumn{4}{|c|}{\textbf{\# FRSs (\textit{c})}}\\ \cline{3-6}
\multicolumn{2}{c|}{} & \textbf{1} & \textbf{2} & \textbf{3} & \textbf{4}\\
\hline
\multirow{10}{*}{\rotatebox[origin=c]{90}{\textbf{\# Attempts (\textit{r})}}} & \textbf{1} & 25.8\% & 8.4\% & 2.7\% & 0.8\%  \\ \cline{2-6}
& \textbf{2} & 20.7\% & 6.0\% & 2.1\% & 0.6\%  \\ \cline{2-6}
& \textbf{3} & 17.9\% & 4.8\% & 1.6\% & 0.4\%  \\ \cline{2-6}
& \textbf{4} & 15.7\% & 4.0\% & 1.5\% & 0.3\%  \\ \cline{2-6}
& \textbf{5} & 13.9\% & 3.7\% & 1.4\% & 0.2\%  \\ \cline{2-6}
& \textbf{6} & 11.3\% & 3.3\% & 1.0\% & 0.1\%  \\ \cline{2-6}
& \textbf{7} & 9.6\% & 3.0\% & 0.7\% & 0.0\% \\ \cline{2-6}
& \textbf{8} & 8.1\% & 2.6\% & 0.6\% & 0.0\% \\ \cline{2-6}
& \textbf{9} & 7.0\% & 1.8\% & 0.3\% & 0.0\% \\ \cline{2-6}
& \textbf{10} & 4.8\% & 0.9\% & 0.0\% & 0.0\%  \\
\hline
\end{tabular}
\end{center}
\end{table}

Finally, as a last element of evaluation, we computed MAP for the morphed images obtained with two of the morphing algorithms used in SOTAMD able to generate the most realistic morphed images: C02 and C05 (see Appendix C in \cite{9246583} for detailed information on the different morphing algorithms used in the generation of SOTAMD dataset). The results are given in Tables~\ref{table:C02} and \ref{table:C05}, respectively. Even in this case, we can argue that the attack potential of the two algorithms is quite different: while the morphed images generated by C02 present a higher attack potential with respect to those generated by C05 to fool one or two different FRSs using up to seven different probe images (i.e., $\mbox{MAP}[1:7,1:2]$), morphed images generated by C05 present a slightly more robust (i.e., $\mbox{MAP}[8:10,1:4]$) and general attack potential (i.e., $\mbox{MAP}[1:10,3:4]$). 

\begin{table}[ht]
\begin{center}
\caption{MAP computed on the SOTAMD subset of images generated with morphing algorithm C02 (expressed as a percentage) \\ 700 images}
\label{table:C02}
\begin{tabular}{|c|c|c|c|c|c|}
\cline{3-6}
\multicolumn{2}{c|}{\multirow{2}{*}{}} & \multicolumn{4}{|c|}{\textbf{\# FRSs (\textit{c})}}\\ \cline{3-6}
\multicolumn{2}{c|}{} & \textbf{1} & \textbf{2} & \textbf{3} & \textbf{4}\\
\hline
\multirow{10}{*}{\rotatebox[origin=c]{90}{\textbf{\# Attempts (\textit{r})}}} & \textbf{1} & 51.6\% & 19.1\% & 5.4\% & 0.9\%  \\ \cline{2-6}
& \textbf{2} & 43.4\% & 13.7\% & 3.0\% & 0.3\%  \\ \cline{2-6}
& \textbf{3} & 37.1\% & 10.0\% & 1.7\% & 0.1\%  \\ \cline{2-6}
& \textbf{4} & 31.6\% & 8.1\% & 0.7\% & 0.0\%  \\ \cline{2-6}
& \textbf{5} & 28.3\% & 6.9\% & 0.6\% & 0.0\% \\ \cline{2-6}
& \textbf{6} & 21.4\% & 5.6\% & 0.3\% & 0.0\% \\ \cline{2-6}
& \textbf{7} & 16.6\% & 4.1\% & 0.1\% & 0.0\% \\ \cline{2-6}
& \textbf{8} & 11.7\% & 3.4\% & 0.0\% & 0.0\% \\ \cline{2-6}
& \textbf{9} & 8.7\% & 2.1\% & 0.0\% & 0.0\% \\ \cline{2-6}
& \textbf{10} & 5.0\% & 1.0\% & 0.0\% & 0.0\% \\
\hline
\end{tabular}
\end{center}
\end{table}

\begin{table}[ht]
\begin{center}
\caption{MAP computed on the SOTAMD subset of images generated with morphing algorithm C05 (expressed as a percentage) \\ 1359 images}
\label{table:C05}
\begin{tabular}{|c|c|c|c|c|c|}
\cline{3-6}
\multicolumn{2}{c|}{\multirow{2}{*}{}} & \multicolumn{4}{|c|}{\textbf{\# FRSs (\textit{c})}}\\ \cline{3-6}
\multicolumn{2}{c|}{} & \textbf{1} & \textbf{2} & \textbf{3} & \textbf{4}\\
\hline
\multirow{10}{*}{\rotatebox[origin=c]{90}{\textbf{\# Attempts (\textit{r})}}} & \textbf{1} & 38.2\% & 16.5\% & 6.3\% & 1.9\%  \\ \cline{2-6}
& \textbf{2} & 32.3\% & 12.2\% & 4.7\% & 1.3\%  \\ \cline{2-6}
& \textbf{3} & 28.7\% & 9.9\% & 3.8\% & 1.0\%  \\ \cline{2-6}
& \textbf{4} & 24.9\% & 7.4\% & 2.6\% & 0.9\% \\ \cline{2-6}
& \textbf{5} & 22.4\% & 5.8\% & 2.1\% & 0.4\% \\ \cline{2-6}
& \textbf{6} & 18.8\% & 4.9\% & 1.6\% & 0.4\% \\ \cline{2-6}
& \textbf{7} & 15.6\% & 4.1\% & 1.0\% & 0.1\% \\ \cline{2-6}
& \textbf{8} & 13.2\% & 3.5\% & 0.9\% & 0.1\% \\ \cline{2-6}
& \textbf{9} & 10.7\% & 2.6\% & 0.2\% & 0.0\% \\ \cline{2-6}
& \textbf{10} & 7.6\% & 1.8\% & 0.1\% & 0.0\% \\
\hline
\end{tabular}
\end{center}
\end{table}

\section{Conclusions}
In this paper we proposed a novel methodology (called MAP) to measure the attack potential of a morphing method constituted with a dataset of morphed images. This methodology is able to overcome the limitations of existing indicators (i.e., MMPMR and FMMPMR). In particular, MAP proved to be useful when it is necessary to simulate a real border control scenario with a variable number of attempts (i.e., multiple probe images acquired at the gate) and multiple face recognition systems (i.e., to simulate ABC gates from different vendors). 

While the vulnerability reporting of a FRS is the primary objective, it was outlined in \cite{8053499} that a conclusion on the pure MMPMR might be misleading, if the FRSs under test have a significantly different recognition performance. This assumption was later verified in the NIST FRVT-MORPH testing \cite{NIST-FRVT-MORPH}. In order to report both the vulnerability and the recognition performance, in \cite{8053499} a Relative Morph Match Rate (RMMR) was proposed, which unifies the vulnerability measure with the recognition accuracy measure. This proposal was - for vulnerability assessment of presentation attacks - recently incorporated in the revision of the international standard ISO/IEC FDIS 30107-3 \cite{ISO_IEC_DIS_30107-3} as Relative Impostor Attack Presentation Accept Rate (RIPAR). In order to avoid an excessive complexity in the description of the MAP, the recognition accuracy measure has not been incorporated in this work; a possible extension will be considered in the future.

Moreover, although MAP has been explicitly introduced to measure the attack potential of morphing against different FRSs, this metric can be considered a general methodology to evaluate in a consistent manner the attack potential of other kind of presentation attacks (e.g., deepfake or make-up attack) not only based on face (e.g., double-identity fingerprints \cite{double_identity_fingerprints}).

The source code for MAP computation is available on Github\footnote{https://github.com/MatteoFerrara/Morphing-Attack-Potential}.

\bibliography{biblio}

\begin{thebibliography}{10}
\providecommand{\url}[1]{#1}
\csname url@samestyle\endcsname
\providecommand{\newblock}{\relax}
\providecommand{\bibinfo}[2]{#2}
\providecommand{\BIBentrySTDinterwordspacing}{\spaceskip=0pt\relax}
\providecommand{\BIBentryALTinterwordstretchfactor}{4}
\providecommand{\BIBentryALTinterwordspacing}{\spaceskip=\fontdimen2\font plus
\BIBentryALTinterwordstretchfactor\fontdimen3\font minus
  \fontdimen4\font\relax}
\providecommand{\BIBforeignlanguage}[2]{{%
\expandafter\ifx\csname l@#1\endcsname\relax
\typeout{** WARNING: IEEEtran.bst: No hyphenation pattern has been}%
\typeout{** loaded for the language `#1'. Using the pattern for}%
\typeout{** the default language instead.}%
\else
\language=\csname l@#1\endcsname
\fi
#2}}
\providecommand{\BIBdecl}{\relax}
\BIBdecl

\bibitem{6996240}
M.~Ferrara, A.~Franco, and D.~Maltoni, ``The magic passport,'' in \emph{IEEE
  International Joint Conference on Biometrics}, 2014, pp. 1--7.

\bibitem{8053499}
U.~Scherhag, A.~Nautsch, C.~Rathgeb, M.~Gomez-Barrero, R.~N.~J. Veldhuis,
  L.~Spreeuwers, M.~Schils, D.~Maltoni, P.~Grother, S.~Marcel, R.~Breithaupt,
  R.~Ramachandra, and C.~Busch, ``Biometric systems under morphing attacks:
  Assessment of morphing techniques and vulnerability reporting,'' in
  \emph{2017 International Conference of the Biometrics Special Interest Group
  (BIOSIG)}, 2017, pp. 1--7.

\bibitem{VenkateshRRB20-0}
\BIBentryALTinterwordspacing
S.~Venkatesh, K.~B. Raja, R.~Ramachandra, and C.~B. 0001, ``On the influence of
  ageing on face morph attacks: Vulnerability and detection,'' in \emph{2020
  IEEE International Joint Conference on Biometrics, IJCB 2020, Houston, TX,
  USA, September 28 - October 1, 2020}.\hskip 1em plus 0.5em minus 0.4em\relax
  IEEE, 2020, pp. 1--10. [Online]. Available:
  \url{https://doi.org/10.1109/IJCB48548.2020.9304856}
\BIBentrySTDinterwordspacing

\bibitem{ISO_IEC_30107-3_2017}
``Information technology — biometric presentation attack detection — part
  3: Testing and reporting,'' September 2017, {I}SO/IEC 30107-3:2017.

\bibitem{9246583}
K.~Raja, M.~Ferrara, A.~Franco, L.~Spreeuwers, I.~Batskos, F.~de~Wit,
  M.~Gomez-Barrero, U.~Scherhag, D.~Fischer, S.~K. Venkatesh, J.~M. Singh,
  G.~Li, L.~Bergeron, S.~Isadskiy, R.~Ramachandra, C.~Rathgeb, D.~Frings,
  U.~Seidel, F.~Knopjes, R.~Veldhuis, D.~Maltoni, and C.~Busch, ``Morphing
  attack detection-database, evaluation platform, and benchmarking,''
  \emph{IEEE Transactions on Information Forensics and Security}, vol.~16, pp.
  4336--4351, 2021.

\bibitem{deepface}
\BIBentryALTinterwordspacing
Deepface website. [Online]. Available:
  \url{https://github.com/serengil/deepface}
\BIBentrySTDinterwordspacing

\bibitem{FrontexGuidelines}
{R\&D Unit}, ``Best practice technical guidelines for automated border control
  ({ABC}) systems,'' FRONTEX, Tech. Rep., September 2015.

\bibitem{ISO_IEC_19795-1}
``Information technology — biometric performance testing and reporting —
  part 1: Principles and framework,'' May 2021, {I}SO/IEC 19795-1:2021.

\bibitem{Ferrara2016}
M.~Ferrara, A.~Franco, and D.~Maltoni, \emph{On the Effects of Image
  Alterations on Face Recognition Accuracy}.\hskip 1em plus 0.5em minus
  0.4em\relax Cham: Springer International Publishing, 2016, pp. 195--222.

\bibitem{NIST-FRVT-MORPH}
\BIBentryALTinterwordspacing
NIST. Frvt morph. [Online]. Available:
  \url{https://pages.nist.gov/frvt/html/frvt\_morph.html}
\BIBentrySTDinterwordspacing

\bibitem{ISO_IEC_DIS_30107-3}
``Information technology — biometric presentation attack detection — part
  3: Testing and reporting,'' Under development, {I}SO/IEC FDIS 30107-3.

\bibitem{double_identity_fingerprints}
R.~Cappelli, M.~Ferrara, and D.~Maltoni, ``On the feasibility of creating
  double-identity fingerprints,'' \emph{IEEE Transactions on Information
  Forensics and Security}, vol.~12, no.~4, pp. 892--900, April 2017.

\end{thebibliography}
\bibliographystyle{IEEEtran}

\section*{Appendix}
The vulnerability metrics presented in Section~\ref{sec:map} with reference to morphed images obtained from two contributing subjects can easily be extended to a generic number of subjects $N_s$, by simply imposing that the metric-specific condition holds for all the subjects involved:
\begin{equation}
    C_{MMPMR}(M) := \bigwedge_{i=1}^{N_s} \big[ m(M,P_i,F)=1 \big]
\end{equation}

\begin{equation}
    C_{\textit{MinMax}-MMPMR}(M) := \bigwedge_{i=1}^{N_s} \big[ mc(M,\mathbb{P}_i,F) \geq 1 \big]
\end{equation}

\begin{equation}
    C_{FMMPMR}(M):= \bigwedge_{i=1}^{N_s} \big[ mc(M,\mathbb{P}_i,F)=|\mathbb{P}_i| \big]
\end{equation}

\begin{equation}
    C_{MVR[r,c]}(M)=\bigwedge_{i=1}^{N_s} \big[ fmc(M,\mathbb{P}_i,\mathbb{F},r)\geq c \big]
\end{equation}

\end{document}